# Chapter 5

# Minimum Jerk Trajectory Generation for Straight and Curved Movements: Mathematical Analysis

*Abdel-Nasser Sharkawy*

## 5.1. Introduction

Jerk is the time derivative of acceleration, and it is an important factor in both suppressing vibration and achieving the high accuracy [1]. The minimum jerk trajectory as presented in [1] minimizes the acceleration changes of the human hand movement from one point to another. In addition, the lowest effort is required by the human hand during the movement. In general, we can say that the minimum jerk trajectory is the free-hand motion.

Study and using of minimum jerk trajectory were presented with many researchers. In [2], the minimum jerk reaching movements of human arm with mechanical constraints at endpoint was studied. Piazzi and Visioli [3] proposed an interval analysis-based approach for finding the global minimum jerk trajectory of the robotic manipulator within a joint space scheme using cubic splines. In our previous papers [4, 5], the velocity of the minimum jerk trajectory was used as the reference velocity and was compared with the actual velocity of the robot end-effector. The resulted error between these two velocities was used for indirectly adjusting the virtual damping and the virtual inertia of the robot admittance controller using the multilayer feedforward neural network. In [6], polynomials were used in conjunction with minimum jerk theory for creating smooth trajectories.

Abdel-Nasser Sharkawy
Mechatronics Engineering, Mechanical Engineering Department,
Faculty of Engineering, South Valley University, Qena 83523, Egypt

In this chapter, the mathematical analysis for the position and the velocity of the minimum jerk trajectory as a function of time is presented. This analysis includes two cases:

- The first case is the unconstrained point-to-point movements of the human hand;

- The second case is the curved point-to-point movements of the human hand.

In addition, a simulation study using Matlab is executed with some examples. Basically, this analysis and study are very important to be used in human-robot interaction, rehabilitation, and haptic applications.

The rest of this chapter is divided as follows. Section 5.2 presents the minimum jerk trajectory for the straight-line segment movements, whereas Section 5.3 presents the minimum jerk trajectory for the curved movements. In Section 5.4, the applications of the MJT are included. Finally, Section 5.5 summarizes the main points of this work.

## 5.2. Minimum Jerk Trajectory for Unconstrained PTP Movement

The MJT is represented by the following function [1],

$$M = \int_0^{t_f} \|\dddot{x}\|^2 \, dt, \quad (5.1)$$

where, $t_f$ is the duration of motion. When the movement of the human hand takes place along a straight-line segment (from initial position to the final position in a given time), the position of the MJT as a function of time is given by [1]:

$$x(t) = x_0 + (x_f - x_0)(6\tau^5 - 15\tau^4 + 10\tau^3),$$
$$y(t) = y_0 + (y_f - y_0)(6\tau^5 - 15\tau^4 + 10\tau^3), \quad (5.2)$$

where, $\tau$ is the normalized time and equal to $t/t_f$, $0 \leq \tau \leq 1$. $x_0, y_0$ are the initial position coordinates at $t = 0$, and $x_f, y_f$ are the final positions coordinates at $t = t_f$.

Assuming the velocity at the beginning and end of the movement is zero ($\dot{x}_0 = \dot{x}_{t_f} = 0, \dot{y}_0 = \dot{y}_{t_f} = 0$), then the velocity of the MJT can be calculated through numeric differentiation as the following equation

$$V_{jerkx} = \dot{x}(t) = x_f\,(30\,\tau^4 - 60\,\tau^3 + 30\,\tau^2),$$

$$V_{jerky} = \dot{y}(t) = y_f\,(30\,\tau^4 - 60\,\tau^3 + 30\,\tau^2), \quad (5.3)$$

In this case, the movement of the human hand can be along $x$ −axis, $y$ −axis, or between $x$ − and $y$ − axes, as shown in Fig. 5.1. Assume, the movement along $x$ −axis, $x_f = 0.3\,m$, $t_f = 1$ seconds, and $t = [0, 1]$ seconds, then the position and the velocity of the MJT are presented using MATLAB in Fig. 5.2.

It should be noted that, if the movement is along $y$ −axis, or between $x$ − and $y$ − axes, the same procedures can be used to obtain the position and velocity diagrams of the MJT.

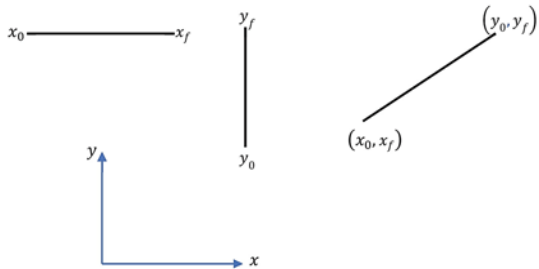

**Fig. 5.1.** The movement of the human hand in straight-line segment along $x$ −axis, $y$ −axis, or between $x$ − and $y$ − axes.

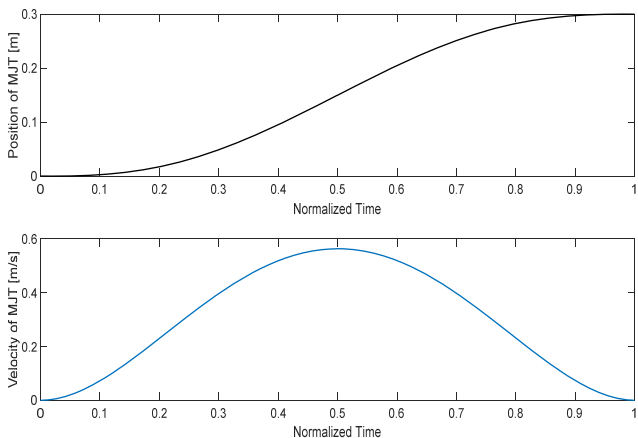

**Fig. 5.2.** The position and the velocity of the MJT. The movement of the human hand is along the $x$ −axis. Normalized time is $\tau = t/t_f$.

In the minimum jerk trajectory, the velocity is high around the middle of the straight-line segment and low, tending to zero, close to the ends where high accuracy is required for accurate positioning of the carried object. Therefore, in this task, high and low velocity motions are combined.

## 5.3. Minimum Jerk Trajectory for Curved PTP Movement

For modelling the curved movement, the human hand in its motion between the end points must pass through a third specified point [1]. Therefore, the movement objective is stated as follows: To generate the smoothest curved motion to bring the human hand from the initial position to the final position in a given time, the human hand must move to the final position through a specified point at an unspecified time. In this case if the location of this specified point with respected to the Cartesian coordinate system is given by the coordinates $(x_1, y_1)$, the equality constraints on the human hand position coordinates $x(t)$ and $y(t)$ at intermediate time $t_1$ are as follows:

$$x(t_1) = x_1, y(t_1) = y_1 \qquad (5.4)$$

This intermediate time $t_1$ where the human hand must pass through is not a priori specified. It will be obtained from procedure as we describe later in the current chapter.

The position component $x(t)$ at all times $t \leq t_1$ [1].

$$x^-(\tau) = \frac{t_f^5}{720} \left( \pi_1 (\tau_1^4 (15\tau^4 - 30\tau^3) + \tau_1^3 (80\tau^3 - 30\tau^4) - 60\tau^3 \tau_1^2 + 30\tau^4 \tau_1 - 6\tau^5) + c_1 (15\tau^4 - 10\tau^3 - 6\tau^5) \right) + x_0 \qquad (5.5)$$

And for times $t \geq t_1$, the expression is as follows

$$x^+(\tau) = x^-(\tau) + \pi_1 \frac{\tau_1^5 (\tau - \tau_1)^5}{120}, \qquad (5.6)$$

where $\tau = \frac{t}{t_f}$ and $\tau_1 = \frac{t_1}{t_f}$. $c_1$ and $\pi_1$ are constants. By substituting $x^+(t_1) = x^-(t_1) = x_1$ in equations (5.5) and (5.6) and solving for $\pi_1$ and $c_1$, then these constants are defined as follows:

$$c_1 = \frac{1}{t_f^5 \tau_1^2 (1-\tau_1)^5} \Big( (x_f - x_0)(300\tau_1^5 - 1200\tau_1^4 + 1600\tau_1^3) + \tau_1^2(-720x_f +$$
$$+120x_1 + 600x_0) + (x_0 - x_1)(300\tau_1 - 200) \Big) = \frac{1}{t_f^5 \tau_1^2 (1-\tau_1)^5} A_1, \quad (5.7)$$

$$\pi_1 = \frac{1}{t_f^5 \tau_1^5 (1-\tau_1)^5} \left( \frac{(x_f - x_0)(120\tau_1^5 - 300\tau_1^4 + 200\tau_1^3) -}{20(x_1 - x_0)} \right)^1 =$$
$$= \frac{1}{t_f^5 \tau_1^5 (1-\tau_1)^5} A_2 \qquad (5.8)$$

The velocity of the minimum jerk trajectory model is given by calculating the numeric differentiation of equations (5.5) and (5.6) as follows:

$$V_{jerkx}^- = \frac{t_f^5}{720} \left( \pi_1 \begin{pmatrix} \tau_1^4(60\tau^3 - 90\tau^2) + \tau_1^3(240\tau^2 - 120\tau^3) - \\ 180\tau^2\tau_1^2 + 120\tau^3\tau_1 - 30\tau^4 \\ +c_1(60\tau^3 - 30\tau^2 - 30\tau^4) \end{pmatrix} + \right), \quad (5.9)$$

$$V_{jerkx}^+ = V_{jerk}^- + 5\pi_1 \frac{\tau_1^5(\tau-\tau_1)^4}{120} = V_{jerk}^- + \pi_1 \frac{\tau_1^5(\tau-\tau_1)^4}{24} \qquad (5.10)$$

**Note:** The same expressions are obtained for $y^-$ and for $y^+$ and $\pi_2$ and $c_2$ replacing $\pi_1$ and $c_1$. In addition, $V_{jerky}^-$ and $V_{jerky}^+$ in the $y$ direction.

The main point now is to find the value of $\tau_1$. The following procedure is performed to find this value.

From Hamiltonian equation presented in [1],

$$\pi_1 u(t_1) + \pi_2 v(t_1) = 0, \qquad (5.11)$$

where $u(t_1) = \dot{x}_1$ and $v(t_1) = \dot{y}_1$.

As known, $x^+(t_1) = x^-(t_1) = x_1$ So $\dot{x}^+(t_1) = \dot{x}^-(t_1) = \dot{x}_1$,

$$y^+(t_1) = y^-(t_1) = y_1 \text{ So } \dot{y}^+(t_1) = \dot{y}^-(t_1) = \dot{y}_1 \quad (5.12)$$

From equation (5.12), then equation (5.11) can be converted to

$$\pi_1 \dot{x}_1 + \pi_2 \dot{y}_1 = 0 \qquad (5.13)$$

---

[1] This is not a matrix. We put the calculations in parentheses like that to present them easier.

From equation (5.5),

$$\dot{x}_1 = \frac{t_f^5}{720}\left(\pi_1 \begin{pmatrix} \tau_1^4(60\tau_1^3 - 90\tau_1^2) + \tau_1^3(240\tau_1^2 - 120\tau_1^3) - 180\tau_1^4 + \\ +120\tau_1^4 - 30\tau_1^4 \\ +c_1\,(60\tau_1^3 - 30\tau_1^2 - 30\tau_1^4) \end{pmatrix} + \right) =$$

$$= \frac{t_f^5}{720}\left(\pi_1(60\tau_1^7 - 90\tau_1^6 + 240\tau_1^5 - 120\tau_1^6 - 90\tau_1^4) + c_1\,(60\tau_1^3 - 30\tau_1^2 - 30\tau_1^4)\right) = \frac{t_f^5}{720}\left(\pi_1(60\tau_1^7 - 210\tau_1^6 + 240\tau_1^5 - 90\tau_1^4) + c_1\,(60\tau_1^3 - 30\tau_1^2 - 30\tau_1^4)\right) = \frac{t_f^5}{720}(\pi_1 A_3 + c_1\,A_4), \qquad (5.14)$$

$$\dot{y}_1 = \frac{t_f^5}{720}\left(\pi_2(\tau_1^4(60\tau_1^3 - 90\tau_1^2) + \tau_1^3(240\tau_1^2 - 120\tau_1^3) - 180\tau_1^4 + 120\tau_1^4 - 30\tau_1^4) + c_2\,(60\tau_1^3 - 30\tau_1^2 - 30\tau_1^4)\right) = \frac{t_f^5}{720}\left(\pi_2(60\tau_1^7 - 210\tau_1^6 + 240\tau_1^5 - 90\tau_1^4) + c_2\,(60\tau_1^3 - 30\tau_1^2 - 30\tau_1^4)\right) = \frac{t_f^5}{720}(\pi_2 A_3 + c_2\,A_4) \qquad (5.15)$$

By substituting equations (5.14) and (5.15) into equation (5.13), then we can get

$$\pi_1 \dot{x}_1 + \pi_2 \dot{y}_1 = 0,$$

$$\pi_1 \left(\frac{t_f^5}{720}(\pi_1 A_3 + c_1\,A_4)\right) + \pi_2 \left(\frac{t_f^5}{720}(\pi_2 A_3 + c_2\,A_4)\right) = 0.0,$$

$$\pi_1(\pi_1 A_3 + c_1\,A_4) + \pi_2(\pi_2 A_3 + c_2\,A_4) = 0.0,$$

$$\pi_1^2 A_3 + \pi_1 c_1\,A_4 + \pi_2^2 A_3 + \pi_2\,c_2\,A_4 = 0.0,$$

$$A_3(\pi_1^2 + \pi_2^2) + A_4(\pi_1 c_1 + \pi_2\,c_2) = 0.0 \qquad (5.16)$$

By substituting the expressions for $\pi_1, \pi_2, c_1,$ and $c_2$ into equation (5.16), then we can get

$$A_3\left(\left(\frac{1}{t_f^5 \tau_1^5(1-\tau_1)^5}A_2\right)^2 + \left(\frac{1}{t_f^5 \tau_1^5(1-\tau_1)^5}A_2'\right)^2\right) +$$

$$+ A_4\begin{pmatrix} \frac{1}{t_f^5 \tau_1^5(1-\tau_1)^5}A_2 \frac{1}{t_f^5 \tau_1^2(1-\tau_1)^5}A_1 + \\ + \frac{1}{t_f^5 \tau_1^5(1-\tau_1)^5}A_2' \frac{1}{t_f^5 \tau_1^2(1-\tau_1)^5}A_1' \end{pmatrix} = 0.0,$$

$$A_3\left(\left(\frac{1}{\tau_1^5}A_2\right)^2 + \left(\frac{1}{\tau_1^5}A_2'\right)^2\right) + A_4\left(\frac{1}{\tau_1^5}A_2\frac{1}{\tau_1^2}A_1 + \frac{1}{\tau_1^5}A_2'\frac{1}{\tau_1^2}A_1'\right) = 0.0,$$

$$A_3\left(\frac{1}{\tau_1^{10}}A_2^2 + \frac{1}{\tau_1^{10}}A_2'^2\right) + A_4\left(\frac{1}{\tau_1^7}A_1A_2 + \frac{1}{\tau_1^7}A_1'A_2'\right) = 0.0,$$

$$A_3A_2^2 + A_3A_2'^2 + \tau_1^3 A_1A_2A_4 + \tau_1^3 A_1'A_2'A_4 = 0.0 \quad (5.17)$$

By substituting the expressions of $A_1, A_2, A_3, A_4, A_1', A_2'$ into equation (5.17) and solving the polynomial equation using MATLAB, we can find the roots of $\tau_1$. From the roots of $\tau_1$, we have only one acceptable root and it is between 0 and 1. The Matlab code for solving this polynomial equation is presented in Appendix 1.

### 5.3.1. Some Examples of Movements

In this subsection, we present four cases; two cases show the movement of the human hand along the $y-$axis, whereas the other two cases show the movement along $x-$axis.

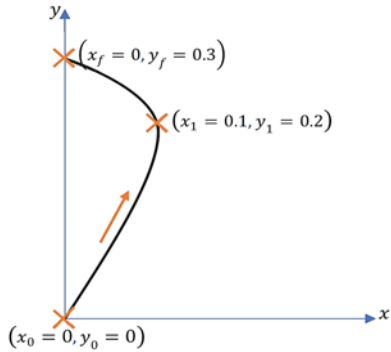

**Fig. 5.3.** The movement of the hand along $y-$axis. $x_f = 0.0\ m$, $y_f = 0.3\ m$, $x_1 = 0.1\ m$, and $y_1 = 0.2\ m$.

**First case:**

Assume the movement of the human hand is along the $y-$axis, $x_0 = 0.0\ m$, $y_0 = 0.0\ m$, $x_f = 0.0\ m$, $y_f = 0.3\ m$, $x_1 = 0.1\ m$, and $y_1 = 0.2\ m$. This curved movement is presented in Fig. 5.3. By

193

solving the polynomial equation presented in (5.17) using MATLAB and function (vpasolve), the only acceptable root between 0 and 1 is $\tau_1 = 0.553$. To plot the position and the velocity of the MJT along $x-$ and $y-$ axes, assume $t_f = 1$ seconds and $t = [0, 1]$ seconds. These variables are presented in Fig. 5.4 using Matlab.

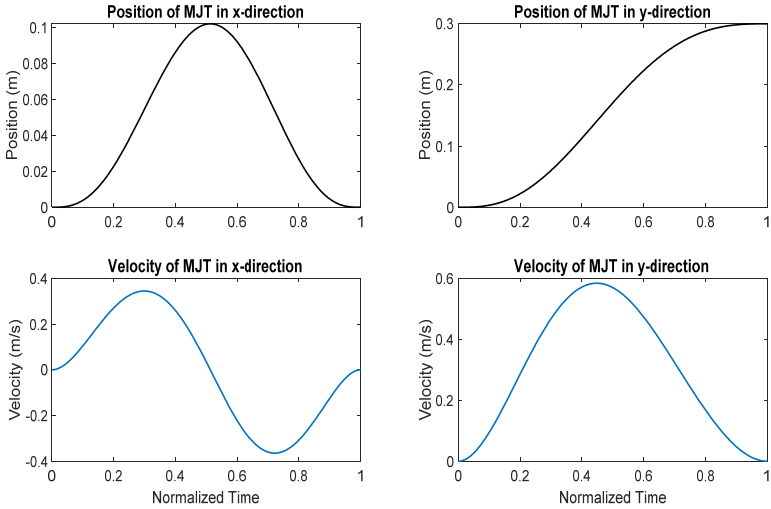

**Fig. 5.4.** The position and the velocity of the MJT during the movement of the hand along $y-$axis. $x_f = 0.0\ m$, $y_f = 0.3\ m$, $x_1 = 0.1\ m$, and $y_1 = 0.2\ m$.

**Second case:**

Assume the movement of the human hand is along the $y-$axis, $x_0 = 0.0\ m$, $y_0 = 0.0\ m$, $x_f = 0.0\ m$, $y_f = 0.4\ m$, $x_1 = 0.1\ m$, and $y_1 = 0.1\ m$. This curved movement is presented in Fig. 5.5. By solving the polynomial equation presented in (5.17) using MATLAB and function (vpasolve), the only acceptable root between 0 and 1 is $\tau_1 = 0.4055$. To plot the position and the velocity of the MJT along $x-$ and $y-$ axes, assume $t_f = 1$ seconds and $t = [0, 1]$ seconds. These variables are presented in Fig. 5.6 using Matlab.

**Third case:**

Assume the movement of the human hand is along the $x-$axis, $x_0 = 0.0\ m$, $y_0 = 0.0\ m$, $x_f = 0.3\ m$, $y_f = 0.0\ m$, $x_1 = 0.2\ m$, and $y_1 = 0.2\ m$. This curved movement is presented in Fig. 5.7. By solving the polynomial equation presented in (5.17) using MATLAB and function (vpasolve), the only acceptable root between 0 and 1 is $\tau_1 = 0.5245$. To plot the position and the velocity of the MJT along $x-$ and $y-$ axes, assume $t_f = 1$ seconds and $t = [0, 1]$ seconds. These variables are presented in Fig. 5.8 using Matlab.

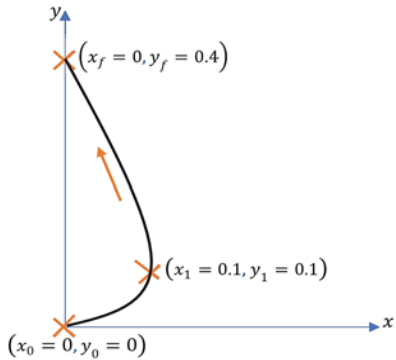

**Fig. 5.5.** The movement of the hand along $y-$axis. $x_f = 0.0\ m$, $y_f = 0.4\ m$, $x_1 = 0.1\ m$, and $y_1 = 0.1\ m$.

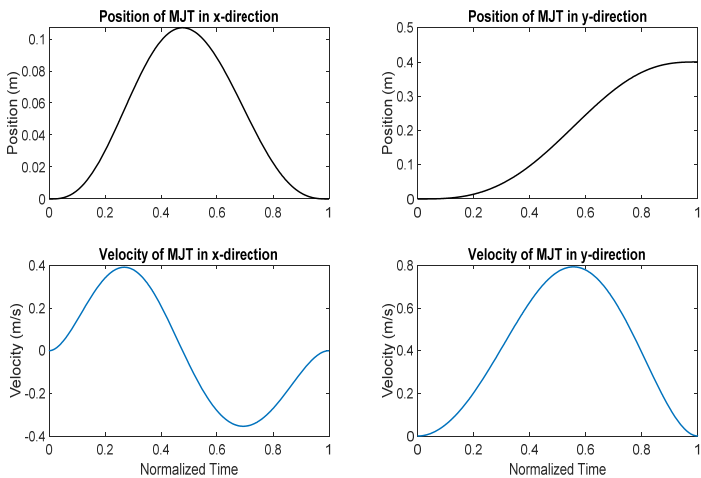

**Fig. 5.6.** The position and the velocity of the MJT during the movement of the hand along $y-$axis. $x_f = 0.0\ m$, $y_f = 0.4\ m$, $x_1 = 0.1\ m$, and $y_1 = 0.1\ m$.

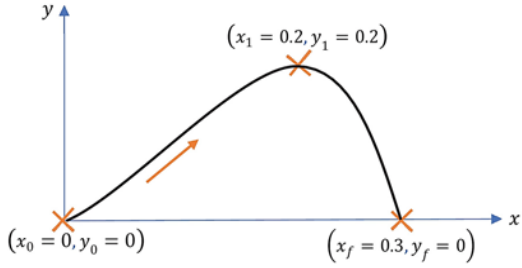

**Fig. 5.7.** The movement of the hand along $x$ −axis. $x_f = 0.3\ m$, $y_f = 0\ m$, $x_1 = 0.2\ m$, and $y_1 = 0.2\ m$.

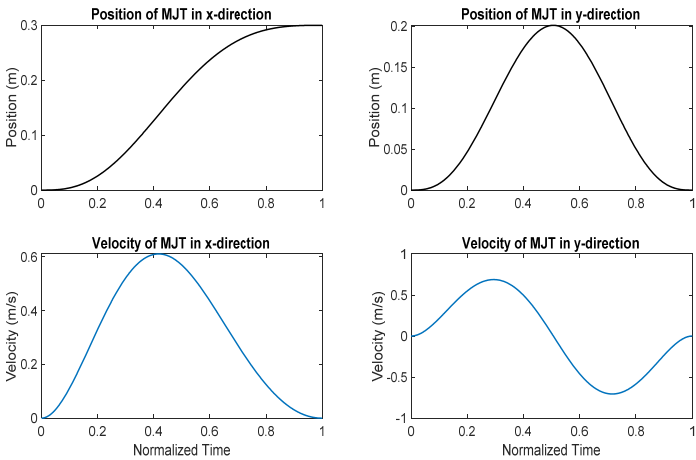

**Fig. 5.8.** The position and the velocity of the MJT during the movement of the hand along $x$ −axis. $x_f = 0.3\ m$, $y_f = 0\ m$, $x_1 = 0.2\ m$, and $y_1 = 0.2\ m$.

**Fourth case:**

Assume the movement of the human hand is along the $x$ −axis, $x_0 = 0.0\ m$, $y_0 = 0.0\ m$, $x_f = 0.6\ m$, $y_f = 0.0\ m$, $x_1 = 0.1\ m$, and $y_1 = 0.3\ m$. This curved movement is presented in Fig. 5.9. By solving the polynomial equation presented in (5.17) using MATLAB and function (vpasolve), the only acceptable root between 0 and 1 is $\tau_1 = 0.4355$. To plot the position and the velocity of the MJT along $x−$ and $y−$ axes, assume $t_f = 2$ seconds and $t = [0,2]$ seconds. These variables are presented in Fig. 5.10 using Matlab.

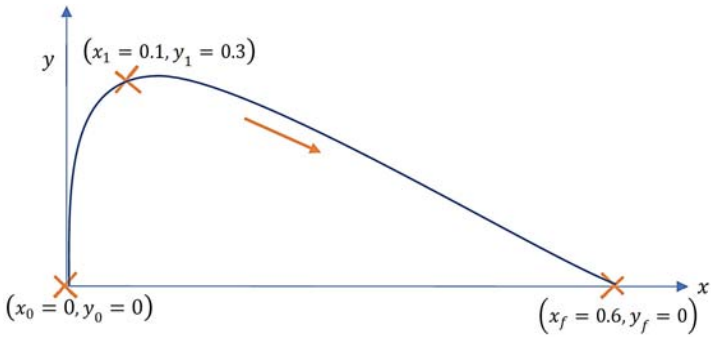

**Fig. 5.9.** The movement of the human hand along $x-$axis. $x_f = 0.6\ m$, $y_f = 0\ m$, $x_1 = 0.1\ m$, and $y_1 = 0.3\ m$.

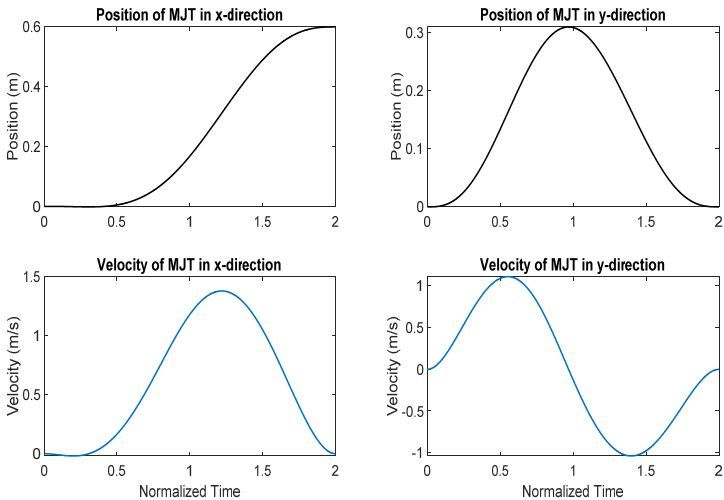

**Fig. 5.10.** The position and the velocity of the MJT during the movement of the hand along $x-$axis. $x_f = 0.6\ m$, $y_f = 0\ m$, $x_1 = 0.1\ m$, and $y_1 = 0.3\ m$.

As presented from Fig. 5.1 to Fig. 5.2 where the straight-line segment movement, and from Fig. 5.4 to Fig. 5.10 where the curved movement, the position and the velocity of the minimum jerk trajectory are very smooth. In general, we can say that the minimum jerk trajectory is the smoothest possible trajectory for the human hand movement.

197

## 5.4. MJT Applications

Minimum jerk trajectory is used with many applications such as control, human-robot interaction, rehabilitation, haptic applications, and mobile robots, etc.

In reactor control application [7], the minimization of the jerk of the power reference trajectory led to the minimization of wear and tear in the control rod drive mechanism. In addition, the tracking of minimum jerk power reference trajectory exhibited a smoother control reactivity profile.

Minimum jerk trajectory was also important for human-robot cooperation. In [4, 5], a neural network system was used for adjusting the virtual damping or the virtual inertia of the robot admittance controller. In this approach, the training of the neural network was indirectly and online using the velocity error between the reference velocity of the minimum jerk trajectory and the actual velocity of the robot. The robot was guided by the human operator to execute a straight-line segment movement.

MJT was also necessary with rehabilitation, haptic applications. In [6], the MJT was intended for retraining the upper limb movements of patients that lose manipulation functions following stroke. In [8], an effective application of the Jerk was proposed for planning trajectory and for controlling the undesirable effects (e.g. patient's hesitations and tremors) in active rehabilitation training using some mechatronic devices.

MJT was used with mobile robots. In [9], a scheme was provided for generating optimal trajectories for the nonholonomic mobile robots. These trajectories minimized the jerk of the trajectory of the center of mass of the robot. In addition, the trajectories were constrained to pass within certain predefined neighborhoods at certain given predefined time instances. MJT was able to ensure physical hardware limitations of the electromagnetic actuators, to prevent undesirable induced vibrations, and to ensure comfort of the passengers.

## 5.5. Conclusions

In this manuscript, the mathematical analysis of the minimum jerk trajectory is presented. In this analysis, the straight-line segment movement of the human hand as well as the curved movement are considered. Simulation study is executed with some examples to plot the position and the velocity of the MJT. The results from this simulation prove that the MJT is the smoothest possible movement of the human hand. In final, MJT is used with many applications such as human-robot interaction, rehabilitation, haptic application, and mobile robots.

## Acknowledgements

The authors would like to thank Prof. Nikos Aspragathos, retired professor, Department of Mechanical Engineering and Aeronautics, University of Patras, Greece, for his support. Furthermore, Prof. Vassilis C. Moulianitis, Professor, Department of Product and Systems Design Engineering, University of the Aegean, Greece, for his help in making the MATLAB code used to plot the position and the velocity of the minimum jerk trajectory.

**Conflict of interest:** The authors declare that they have no conflict of interest.

## Appendix 1. The MATLAB Code for Solving the Polynomial Equation (5.17)

The MATLAB code for solving the polynomial equation (5.17) to find the roots of $\tau_1$ is presented as follows:

```
clc
clear all;
syms t1 tf xf yf x1 y1 float
t1 = sym('t1', 'real');
tf = sym('tf', 'real');
xf = sym('xf', 'real');
yf = sym('yf', 'real');
x1 = sym('x1', 'real');
y1 = sym('y1', 'real');
xf = 0.6
yf = 0.0
x1 = 0.1
y1 = 0.3
%%
A2 = expand(xf*(120*t1^5 - 300*t1^4 + 200*t1^3) - 20*x1)
A22 = expand(yf*(120*t1^5 - 300*t1^4 + 200*t1^3) - 20*y1)
```

```
A1 = expand(xf*(300*t1^5 - 1200*t1^4 + 1600*t1^3)
+ t1^2*(-720*xf + 120*x1) - x1*(300*t1 - 200))
A11 = expand(yf*(300*t1^5 - 1200*t1^4 + 1600*t1^3)
+ t1^2*(-720*yf + 120*y1) - y1*(300*t1 - 200))
A3 = 60*t1^7 - 210*t1^6 + 240*t1^5 - 90*t1^4
A4 = 60*t1^3 - 30*t1^2 - 30*t1^4
Result = expand(A3*A2*A2 + A3*A22*A22 +
(t1^3)*A1*A2*A4 + (t1^3)*A11*A22*A4)
%numerical solution
solvet = vpasolve(Result, t1)
```

`t1` in MATLAB code is the same with $\tau_1$.